\documentclass{article}

\usepackage[preprint,nonatbib]{neurips_2021}

\usepackage[normalem]{ulem}
\usepackage{colortbl}
\usepackage[utf8]{inputenc} 
\usepackage[T1]{fontenc}    
\usepackage{hyperref}       
\usepackage{url}            
\usepackage{booktabs}       
\usepackage{amsfonts}       
\usepackage{nicefrac}       
\usepackage{microtype}      
\usepackage{xcolor}         
\usepackage{subfigure}
\usepackage{graphicx}
\usepackage{amsmath}
\usepackage{wrapfig}
\usepackage{graphbox}
\usepackage{multirow}
\usepackage{longtable}
\usepackage{todonotes}
\usepackage{color}
\usepackage{titlesec}
\usepackage{appendix}
\usepackage{algorithm} 
\usepackage{algpseudocode}

\title{ChemRL-GEM: Geometry Enhanced Molecular Representation Learning for Property Prediction}

\author{%
 Xiaomin Fang$^1$\thanks{Equal contribution.}, Lihang Liu$^1$\footnotemark[1], Jieqiong Lei$^2$, Donglong He$^1$ \\
 \textbf{Shanzhuo Zhang$^1$, Jingbo Zhou$^1$, Fan Wang$^1$\thanks{Corresponding author.}, Hua Wu$^1$, Haifeng Wang$^1$} \\
 $^1$Baidu Inc., China  \\
 $^2$Shenzhen International Graduate School, Tsinghua University \\
 \texttt{\{fangxiaomin01, liulihang\}@baidu.com} \\ \texttt{leijq19@mails.tsinghua.edu.cn} \\
 \texttt{\{hedonglong, zhangshanzhuo, zhoujingbo\}@baidu.com} \\
 \texttt{\{wang.fan, wu\_hua, wanghaifeng\}@baidu.com}
}

\begin{document}
\setlength{\belowcaptionskip}{-1em}

\maketitle

\begin{abstract}

    Effective molecular representation learning is of great importance to facilitate molecular property prediction, which is a fundamental task for the drug and material industry.
    Recent advances in graph neural networks (GNNs) have shown great promise in applying GNNs for molecular representation learning. Moreover, a few recent studies have also demonstrated successful applications of self-supervised learning methods to pre-train the GNNs to overcome the problem of insufficient labeled molecules. However, existing GNNs and pre-training strategies usually treat molecules as topological graph data without fully utilizing the molecular geometry information. Whereas, the three-dimensional (3D) spatial structure of a molecule, a.k.a molecular geometry, is one of the most critical factors for determining molecular physical, chemical, and biological properties. To this end, we propose a novel \textbf{G}eometry \textbf{E}nhanced \textbf{M}olecular representation learning method (GEM) for \textbf{Chem}ical \textbf{R}epresentation \textbf{L}earning (ChemRL). At first, we design a geometry-based GNN architecture that simultaneously models atoms, bonds, and bond angles in a molecule. To be specific, we devised double graphs for a molecule: The first one encodes the atom-bond relations; The second one encodes bond-angle relations. 
    Moreover, on top of the devised GNN architecture, we propose several novel geometry-level self-supervised learning strategies to learn spatial knowledge by utilizing the local and global molecular 3D structures.
    We compare ChemRL-GEM with various state-of-the-art (SOTA) baselines on different molecular benchmarks and exhibit that ChemRL-GEM can significantly outperform all baselines in both regression and classification tasks. For example, the experimental results show an overall improvement of $8.8\%$ on average compared to SOTA baselines on the regression tasks, demonstrating the superiority of the proposed method.
\end{abstract}

\section{Introduction}



Molecular property prediction has been widely considered as one of the most critical tasks in computational drug and materials discovery, since many methods rely on predicted molecular properties to evaluate, select and generate molecules \cite{shen2020molecular,wieder2020compact}. 
With the development of deep neural networks (DNNs), molecular representation learning exhibits a great advantage over feature engineering-based methods, which has attracted increasing research attention to tackle the molecular property prediction problem.

Recently, graph neural networks (GNNs) for molecular representation learning have become an emerging research area, which regard the topology of atoms and bonds as a graph, and propagate messages of each element to its neighbors \cite{DBLP:journals/corr/abs-2004-08919,DBLP:conf/nips/RongBXX0HH20,DBLP:journals/corr/abs-1909-00259,DBLP:conf/icdm/ShuiK20}.
However, one major obstacle to hinder the successful application of GNNs (and DNNs) in molecule property prediction is the scarity of labeled data, which is also a common research challenge in Natural Language Processing (NLP) \cite{DBLP:conf/naacl/DevlinCLT19,DBLP:journals/corr/abs-2006-03654} and Computer Vision (CV) \cite{DBLP:journals/corr/DoerschGE15,gidaris2018unsupervised} communities.
Inspired by the success of self-supervised learning, recent studies \cite{DBLP:conf/iclr/HuLGZLPL20,DBLP:conf/nips/RongBXX0HH20} start to utilize large-scale molecules with self-supervised 
methodology to pre-train the molecular representation, which have achieved
substantial improvements.

\begin{wrapfigure}{r}{0.42\textwidth}
\begin{center}
\includegraphics[width=0.42\textwidth]{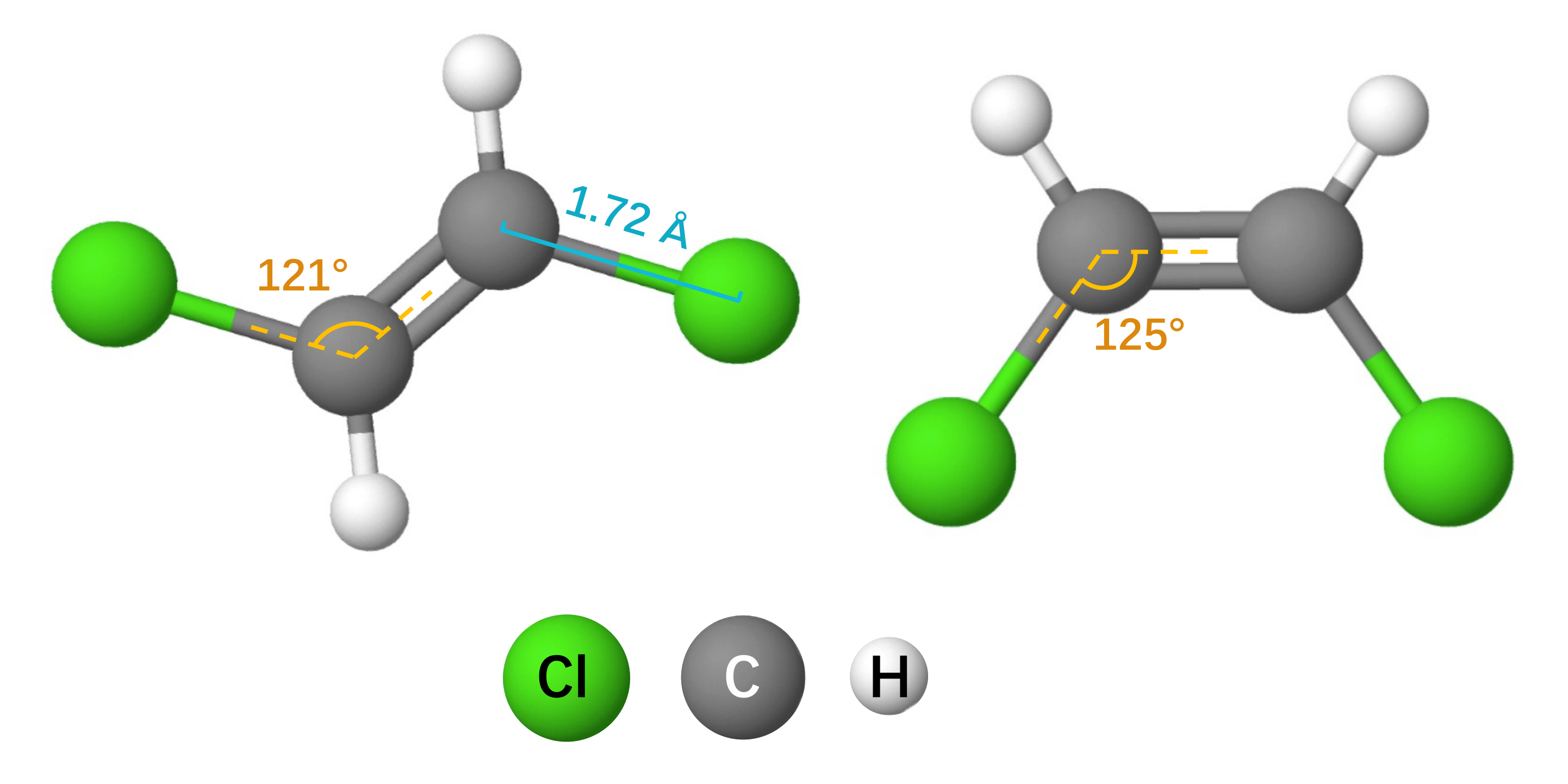}
\end{center}
\caption{Comparison between two molecules (cis-1,2-DCE and trans-1,2-DCE) with the same topology but different geometries. The two chlorine atoms are on the different sides in the left molecule, while the same sides in the right molecule.}\label{fig:geometries1}
\end{wrapfigure}

However, existing representation methods based on GNNs only encode the topology information of the molecules, neglecting the molecular geometry information, i.e. the three-dimensional (3D) spatial structure of a molecule. Although some works \cite{DBLP:conf/nips/SchuttKFCTM17,DBLP:journals/corr/abs-2102-07933} take the atomic distance into edge features to consider partial geometry information, they fail to thoroughly model the geometry information, such as bond-angle information. As a result, existing graph-based molecular representations usually can not distinguish molecules with the same topology but different geometries (i.e., Geometric Isomerism \cite{farrell1995effects}). However, such information plays an important role in determining molecules' physical, chemical, and biological activities. For example, due to the geometrical difference, water solubility (a critical metric of drug-likeness) of the two molecules illustrated in Figure~\ref{fig:geometries1} are different, even though they have the same topology. Cis-Platin and trans-Platin are another example that have the same topology but different geometries: cis-Platin is a popular chemotherapy drug used to treat a number of cancers, whereas trans-Platin has no cytotoxic activity \cite{peleg2002interactions}.



\par As for the self-supervised learning methods, previous works typically do masking and predicting in nodes, edges or contexts in the topology \cite{DBLP:conf/iclr/HuLGZLPL20,DBLP:conf/nips/RongBXX0HH20}. We argue that these tasks only enable the model to learn superficial chemical laws such as ``which atom/group could be connected to a double bond'', but lack the ability to learn the geometry knowledge, such as ``the difference between the bond angles of the two molecules'' shown in Figure~\ref{fig:geometries1}. It is also desirable to propose a delicately designed self-supervised learning methods to fully capture the intrinsic differences in atomic interaction forces due to different molecular geometries.


To solve these problems, we propose a novel \textbf{G}eometry \textbf{E}nhanced \textbf{M}olecular representation learning method (GEM)\footnote{https://github.com/PaddlePaddle/PaddleHelix/tree/dev/apps/pretrained\_compound/ChemRL/GEM} for \textbf{Chem}ical \textbf{R}epresentation \textbf{L}earning (ChemRL). Firstly, to make the message passing sensitive to geometries, we model the atoms, bonds, and bond angles simultaneously by designing a geometry-based GNN architecture (GeoGNN), which contains two graphs: the first encodes the atom-bond relations; the second encodes the bond-angle relations. Secondly, we pre-train the GeoGNN to learn the knowledge of both the chemical laws and the geometries by designing various geometry-level self-supervised learning tasks.
To verify the effectiveness of the proposed ChemRL-GEM, we compared it with several SOTA baselines on a dozen of molecular property prediction benchmarks. {Our exhaustive experimental study demonstrates the superiority of ChemRL-GEM.}

Our contributions can be summarized as follows:
\begin{itemize}
    \setlength{\itemsep}{0pt}
    \item We propose a geometry-based graph neural network, GeoGNN, to encode both the topology and the geometry information of molecules.
    \item We introduce multiple geometry-level self-supervised learning tasks to learn the molecular 3D spatial knowledge in addition to other self-supervised learning tasks.
    \item We evaluated ChemRL-GEM thoroughly on various molecular property prediction datasets. Experimental results demonstrate that ChemRL-GEM significantly outperforms competitive baselines on multiple benchmarks.
\end{itemize}

\section{Preliminaries}

\subsection{Graph-based Molecular Representation}
A molecule consists of atoms, and the neighboring atoms are connected by the chemical bonds, which can be naturally represented by a graph $G=(\mathcal{V},\mathcal{E})$, where $\mathcal{V}$ is a node set and $\mathcal{E}$ is an edge set. An atom in the molecule is regarded as a node $v\in \mathcal{V}$ and a chemical bond in the molecule is regarded as an edge $(u, v) \in \mathcal{E}$ connecting atoms $u$ and $v$.

Graph neural networks (GNNs) can be seen as message passing
neural networks \cite{DBLP:conf/icml/GilmerSRVD17}, which is useful for predicting molecular properties. Following the definitions of the previous GNNs \cite{DBLP:conf/iclr/XuHLJ19}, the features of a node $v$ are represented by $x_v$ and the features of an edge $(u, v)$ are represented by $x_{uv}$. Taking node features, edge features and the graph structure as inputs, a GNN learns the representation vectors of the nodes and the entire graph, where the representation vector of a node $v$ is denoted by $h_v$ and the representation vector of the entire graph is denoted by $h_G$. A GNN iteratively updates a node's representation vector by aggregating the messages from the node's neighbors. Given a node $v$, its representation vector $h_v^{(k)}$ at the $k$-th iteration is formalized by
\begin{equation}
    \begin{split}
        a_v^{(k)} &= \textit{AGGREGATE}^{(k)}(\{ (h_v^{(k-1)}, h_u^{(k-1)}, x_{uv} | u \in \mathcal{N}(v)\}), \\
        h_v^{(k)} &= \textit{COMBINE}^{(k)}(h_v^{(k-1)}, a_v^{(k)}).
    \end{split}
    \label{eq:gnn}
\end{equation}
where $\mathcal{N}(v)$ is the set of neighbors of node $v$, $\textit{AGGREGATE}^{(k)}$ is the aggregation function for aggregating messages from a node's neighborhood, and $\textit{COMBINE}^{(k)}$ is the update function for updating the node representation. We initialize $h_v^{(0)}$ by the feature vector of node $v$, i.e., $h_v^{(0)}=x_v$.

$\textit{READOUT}$ function is introduced to integrate the nodes' representation vectors at the final iteration so as to gain the graph's representation vector $h_G$, which is formalized as
\begin{equation}
    h_G=\textit{READOUT}({h_v^{(K)}|v \in \mathcal{V}}),
\end{equation}
where $K$ is the number of iterations. In most cases, $\textit{READOUT}$ is a permutation invariant pooling function, such as summation and maximization.
The graph's representation vector $h_G$ can then be used for downstream task predictions.

\subsection{Pre-training Methods for GNNs}
In the molecular representation learning community, recently several works \cite{DBLP:conf/iclr/SunHV020,DBLP:conf/iclr/HuLGZLPL20,DBLP:conf/nips/RongBXX0HH20} have explored the power of self-supervised learning to improve the generalization ability of GNN models on downstream tasks. They mainly focus on two kinds of self-supervised learning tasks: the node-level (edge-level) tasks and the graph-level tasks.

The node-level self-supervised learning tasks are devised to capture the local domain knowledge. For example, some studies randomly mask a portion of nodes or sub-graphs and then predict their properties by the node/edge representation. The graph-level self-supervised learning tasks are used to capture the global information, like predicting the graph properties by the graph representation. Usually, the graph properties are domain-specific knowledge, such as experimental results from biochemical assays or the existence of molecular functional groups.

\section{The ChemRL-GEM Framework}
\begin{figure}[t]
\centering
  \includegraphics[width=0.9\textwidth]{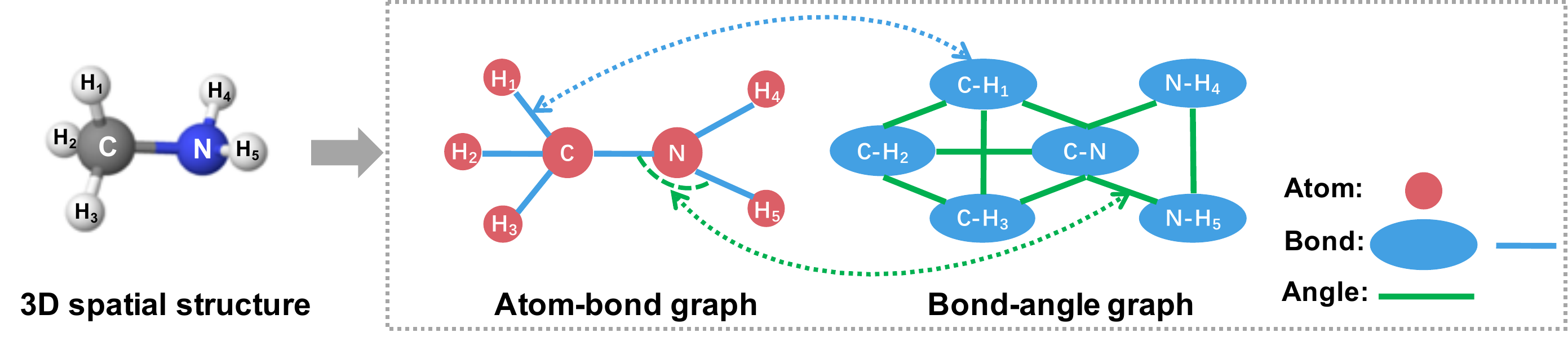}
  \vspace{-0.5em}
  
\caption{Illustration of the atom-bond graph and the bond-angle graph. The left figure shows the structure of Methanamine in the 3D space. We can easily encode its geometry information with the help of the atom-bond graph that describes the relations between atoms and bonds and the bond-angle graph that describes the relations between bonds and bond angles.}
\label{fig:double_graph}
\end{figure}

This section introduces the details of our proposed \textbf{G}eometry \textbf{E}nhanced \textbf{M}olecular representation learning method (GEM) for \textbf{Chem}ical \textbf{R}epresentation \textbf{L}earning, which includes two parts: a novel geometry-based GNN and various geometry-level self-supervised learning tasks.

\subsection{Geometry-based Graph Neural Network}
\label{Sec:geo_gnn}

We propose a \textbf{Geo}metry-based \textbf{G}raph \textbf{N}eural \textbf{N}etwork (GeoGNN) that encodes the molecular geometries by modeling the atom-bond-angle relations, while traditional GNNs only consider the relations between atoms and bonds.


For a molecule, we denote the atom set as $\mathcal{V}$, the bond set as $\mathcal{E}$, and the bond angle set as $\mathcal{A}$. We introduce atom-bond graph $G$, and bond-angle graph $H$ for each molecule, as illustrated in Figure~\ref{fig:double_graph}. The atom-bond graph is defined as $G=(\mathcal{V}, \mathcal{E})$, where atom $u \in \mathcal{V}$ is regarded as the node of $G$ and bond $(u,v) \in \mathcal{E}$ as the edge of $G$, connecting atom $u$ and atom $v$. Similarly, the bond-angle graph is defined as $H=(\mathcal{E}, \mathcal{A})$, where bond $(u,v) \in \mathcal{E}$ is regarded as the node of $H$ and bond angle $(u,v,w) \in \mathcal{A}$ as the edge of $H$, connecting bond $(u,v)$ and bond $(v,w)$. We use $x_u$ as the initial features of atom $u$, $x_{uv}$ as the initial features of bond $(u,v)$ and $x_{uvw}$ as the initial features of bond angle $(u,v,w)$.
The atom-bond graph $G$ and the bond-angle graph $H$, as well as atom features, bond features and bond angle features are taken as the inputs of GeoGNN.
\begin{wrapfigure}{r}{0.47\textwidth}
\centering
\includegraphics[width=0.47\textwidth]{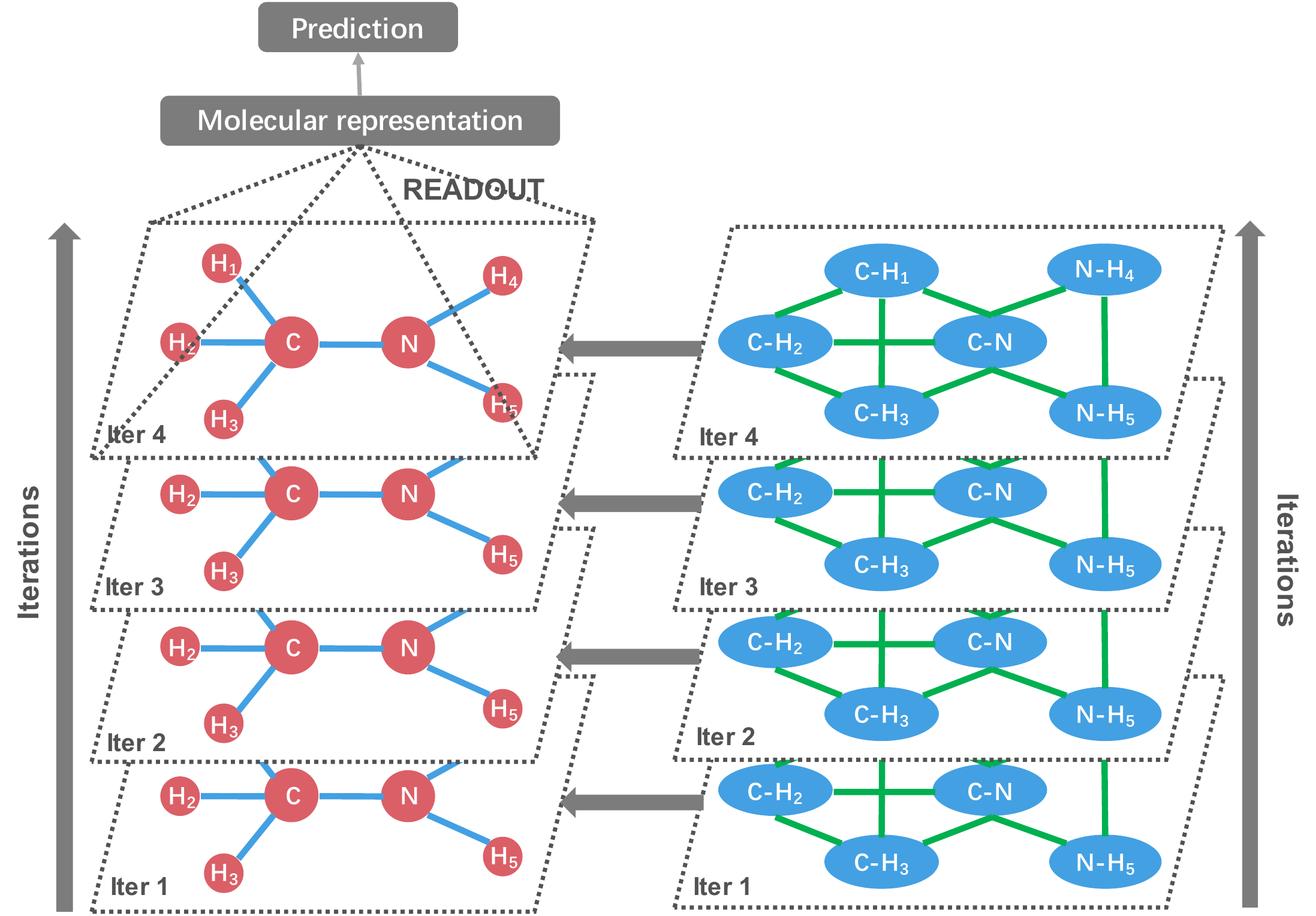}
\caption{Overall architecture of GeoGNN}
\label{fig:geognn_predict}
\end{wrapfigure}

GeoGNN learns the representation vectors of atoms and bonds iteratively. For the $k$-th iteration, the representation vectors of atom $u$ and bond $(u,v)$ are denoted by $h_u$ and $h_{uv}$, respectively. We initialize $h_u^{(0)}=x_u$ and $h_{uv}^{(0)}=x_{uv}$. In order to connect the atom-bond graph $G$ and bond-angle graph $H$, the representation vectors of the bonds as the communication links between $G$ and $H$, as shown in Figure~\ref{fig:geognn_predict}. More concretely, the bonds' representation vectors are learned by aggregating the messages from the neighboring bonds and corresponding bond angles in the bond-angle graph $H$. Then, the learned bond representation vectors are taken as edge features of the atom-bond graph $G$ and help to learn the atoms' representation vectors.

Given bond $(u,v)$, its representation vector $h_{uv}^{(k)}$ at the $k$-th iteration is formalized by
\begin{equation}
    \begin{split}
        a_\textit{uv}^{(k)} = & \textit{AGGREGATE}_\textit{bond-angle}^{(k)}( \{ (h_{uv}^{(k-1)}, h_{uw}^{(k-1)}, x_{wuv}): w \in \mathcal{N}(u)\} \\
        & \qquad\qquad\qquad\qquad\qquad\quad \cup \{ (h_{uv}^{(k-1)}, h_{vw}^{(k-1)}, x_{uvw}): w \in \mathcal{N}(v)\}), \\
        h_{uv}^{(k)} = & \textit{COMBINE}_\textit{bond-angle}^{(k)}(h_{uv}^{(k-1)}, a_{uv}^{(k)}).
    \end{split}
\end{equation}
Here, $\mathcal{N}(u)$ and $\mathcal{N}(v)$ denote the neighboring atoms of atom $u$ and atom $v$, respectively. $\{(u,w):w \in \mathcal{N}(u)\} \cup \{(v,w):w \in \mathcal{N}(v)\}$ are the neighboring bonds of bond $(u,v)$. $\textit{AGGREGATE}_\textit{bond-angle}$ is the message aggregation function, and $\textit{COMBINE}_\textit{bond-angle}$ is the update function for bond-angle graph $H$. In this way, the information from the neighboring bonds and the corresponding bond angles is aggregated into $a_{uv}^{(k)}$. Then, the representation vector of bond $(u,v)$ is updated according to the aggregated information. With the learned representation vectors of the bonds from bond-angle graph $\mathcal{H}$, given an atom $u$, its representation vector $h_u^{(k)}$ at the $k$-th iteration can be formalized as
\begin{equation}
    \begin{split}
        a_u^{(k)} &= \textit{AGGREGATE}_\textit{atom-bond}^{(k)}(\{ (h_u^{(k-1)}, h_v^{(k-1)}, h_{uv}^{(k-1)}): v \in \mathcal{N}(u)\}), \\
        h_u^{(k)} &= \textit{COMBINE}_\textit{atom-bond}^{(k)}(h_u^{(k-1)}, a_u^{(k)}).
    \end{split}
\end{equation}
Similarly, $\mathcal{N}(u)$ denotes the neighboring atoms of atom $u$, $\textit{AGGREGATE}_\textit{atom-bond}$ is the message aggregation function for atom-bond graph $G$, and $\textit{COMBINE}_\textit{atom-bond}$ is the update function. For atom $u$, messages are aggregated from the neighboring atoms and the corresponding bonds. Note that, the messages of the bonds are learned from the bond-angle graph $H$. Then, the aggregated messages update the representation vector of atom $u$.

The representation vectors of the atoms at the final iteration are integrated to gain the molecular representation vector $h_G$ by the $\textit{READOUT}$ function, which is formalized as
\begin{equation}
    h_G=\textit{READOUT}({h_u^{(K)}|u \in \mathcal{V}}),
\end{equation}
where $K$ is the number of iterations. The molecule's representation vector $h_G$ is used to predict the molecular properties.

\subsection{Geometry-level Self-supervised Learning Tasks}
To further boost the generalization ability of GeoGNN, we propose three geometry-level self-supervised learning tasks to pre-train GeoGNN: 1) the bond lengths prediction; 2) the bond angles prediction; 3) the atomic distance matrices prediction. The bond lengths and bond angles describe the local spatial structures, while the atomic distance matrices describe the global spatial structures.

\begin{figure}[t]
\centering
\subfigure[Predict bond lengths]{
    \includegraphics[width=0.29\textwidth]{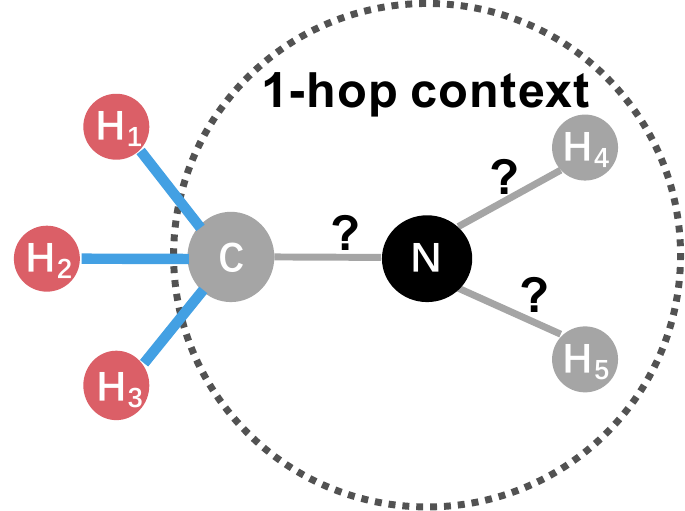}
    \label{fig:bond_length}
    }
\subfigure[Predict bond angles]{
    \includegraphics[width=0.42\textwidth]{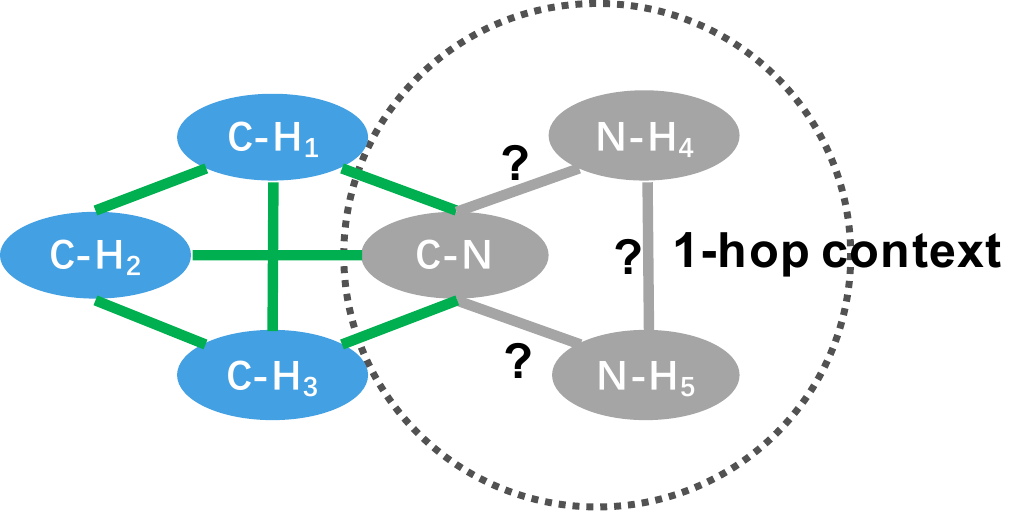}
    \label{fig:bond_angle}
    }
\subfigure[Predict atomic distances]{
    \includegraphics[width=0.24\textwidth]{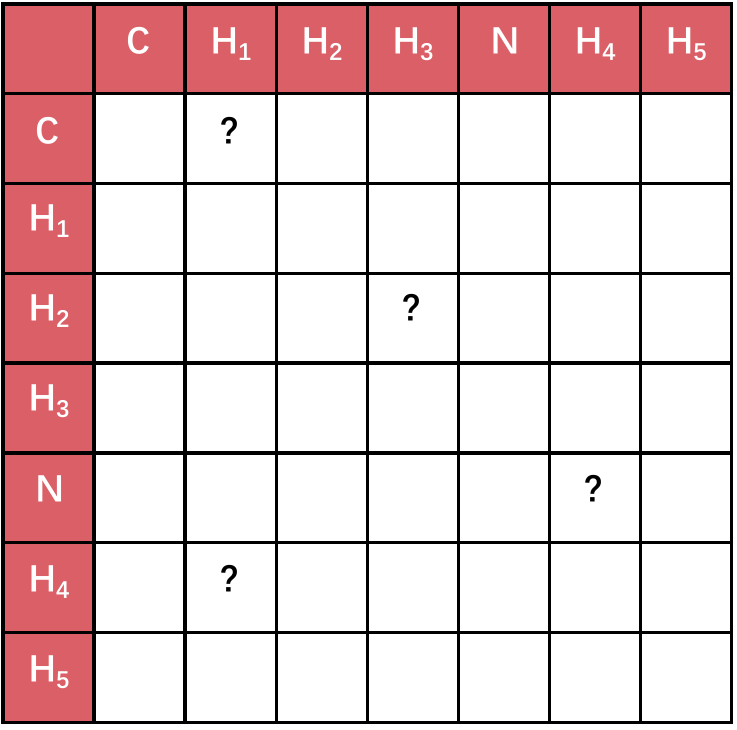}
    \label{fig:atomic_distance}
    }
\label{fig:pretrain}
\caption{Demonstration of geometry-level self-supervised learning tasks. (The black circle represents the selected atom. The gray circles and lines represent the neighboring masked atoms, bonds, and bond angles.)}
\vspace{-0.5em}
\end{figure}

\subsubsection{Local Spatial Structures}
The bond lengths and the bond angles are the most important molecular geometrical parameters. The bond length is the average distance between two joint atoms in a molecule, reflecting the bond strength between the atoms, while the bond angle is the angle connecting two consecutive bonds, including three atoms, describing the local spatial structure of a molecule. 

In order to learn the local spatial structures, we construct self-supervised learning tasks that predict the bond lengths and bond angles. Firstly, for a molecule, we randomly select 15\% of atoms. For each selected atom, we extract 1-hop neighboring atoms and bonds, as well as the bond angles formed by that selected atom. Secondly, we mask the features of these atoms, bonds, and bond angles in the 1-hop context. The representation vectors of the extracted atoms and bonds at the final iteration of GeoGNN are used to predict the extracted bond lengths and bond angles. Figure~\ref{fig:bond_length} and Figure~\ref{fig:bond_angle} show the self-supervised learning tasks based on bond lengths and bond angles. More concretely, for a selected atom $v$, the loss functions of the self-supervised tasks of local geometry information are defined as follow
\begin{equation}
\begin{split}
    L_\textit{length}(\mathcal{E})=&\frac{1}{|\mathcal{E}|}\sum_{(u,v)\in\mathcal{E}}(f_\textit{length}(h_u^{(K)}, h_v^{(K)}) - l_{uv})^2; \\
    L_\textit{angle}(\mathcal{A})=&\frac{1}{|\mathcal{A}|}\sum_{(u,v,w)\in\mathcal{A}}(f_\textit{angle}(h_u^{(K)}, h_v^{(K)}, h_w^{(K)}) - \phi_{uvw})^2.
\end{split}
\label{eq:bond}
\end{equation}
where $f_\textit{length}(\cdot)$ is the network predicting the bond lengths, and $f_\textit{angle}(\cdot)$ is the network predicting the bond angles. $l_{uv}$ denotes the length of the bond connecting atom $u$ and atom $v$ and $\phi_{uvw}$ denotes the degree of the bond angle connecting bonds $(u,v)$ and $(v,w)$. The task of predicting the local spatial structures can be seen as a node-level self-supervised learning task.

\subsubsection{Global Spatial Structures}
Except for the tasks for learning local spatial structures, we also design the atomic distance matrices prediction task for learning the global molecular geometry. We construct the atomic distance matrix for each molecule based on the 3D coordinates of the atoms. Then, we predict the elements in the distance matrix, shown in Figure~\ref{fig:atomic_distance}. We use $d_{uv}$ to denote the distance between two atoms $u$ and $v$ in the molecule.
Note that, for two molecules with the same topological structures, the spatial distances between the corresponding atoms could vary greatly. Thus, for a molecule, rather than take predicting atomic distance matrix as a regression problem, we take it as a multi-class classification problem by discretizing the atomic distances. The loss function is defined as
\begin{equation}
    L_\textit{distance}(\mathcal{V})=\frac{1}{|\mathcal{V}|^2}\sum_{u,v \in \mathcal{V}}-\textit{bin}^T(d_{uv}) \cdot \textit{log}(f_\textit{distance}(h_u^{(K)}, h_v^{(K)})),
\label{eq:distance_matrix}
\end{equation}
where $f_\textit{distance}(\cdot)$ is the network predicting the distribution of atomic distances, the $\textit{bin}(\cdot)$ function is used to discretize the atomic distance $d_{uv}$ into a one-hot vector, and $\textit{log}(\cdot)$ is the logarithmic function. The task predicting the bond lengths can be seen as a special case of the task predicting the atomic distances. The former focuses more on the accurate local spatial structures, while the latter focuses more on the distribution of the global spatial structures. To pre-train GeoGNN, we consider both the local spatial structures and global spatial structures for each molecule by summing up the loss functions defined in Equation~\ref{eq:bond} and Equation~\ref{eq:distance_matrix}.

\section{Experiments}
To thoroughly evaluate the performance of ChemRL-GEM, we compare it with multiple state-of-the-art (SOTA) methods on 12 benchmark datasets from MoleculeNet \cite{DBLP:journals/corr/WuRFGGPLP17} with various molecular property prediction tasks, such as physical, chemical, and biophysics mechanics. The source codes for the experiments will be released to ensure reproducibility when the paper is published.

\subsection{Pre-training Settings}
\textbf{Datasets.}
We use 20 million unlabelled molecules sampled from Zinc15 \cite{DBLP:journals/jcisd/SterlingI15}, a public access database that contains purchasable “drug-like” compounds, to pre-train GeoGNN. We randomly sample 90\% of the molecules for training and the remaining for evaluation.

\textbf{Self-supervised Learning Task Settings.}
We utilize the geometry-level and graph-level tasks to pre-train GeoGNN. For the geometry-level tasks, we utilize the Merck molecular force field \cite{DBLP:journals/jcc/Halgren96} function from the RDKit\footnote{http://www.rdkit.org} package, an open-source cheminformatics toolkit \cite{landrum2006rdkit}, to obtain the simulated 3D coordinates of the atoms in the molecules. The geometric features of the molecule, including bond lengths, bond angles and atomic distance matrices, are calculated by the simulated 3D coordinates.
For the graph-level tasks, we predict two kinds of molecular fingerprints:
1) The Molecular ACCess System (MACCS) key \cite{DBLP:journals/jcisd/DurantLHN02}; 2) The extended-connectivity fingerprint (ECFP) \cite{DBLP:journals/jcisd/RogersH10}.

\subsection{Molecular Property Prediction Settings}
\textbf{Datasets and Splitting Method.}
We conduct experiments on multiple molecular benchmarks from the MoleculeNet \cite{DBLP:journals/corr/WuRFGGPLP17}, including both classification and regression tasks \cite{subramanian2016computational,DBLP:journals/jcisd/MartinsTPF12,richard2016toxcast,gayvert2016data,huang2017editorial,DBLP:journals/nar/KuhnLJB16}. Following the previous work \cite{DBLP:conf/iclr/HuLGZLPL20}, we split all the 12 datasets with scaffold split \cite{ramsundar2019deep}, which splits molecules according to the their scaffold (molecular substructure). Rather than splitting the dataset randomly, scaffold split is a more challenging splitting method, which can better evaluate the generalization ability of the models on out-of-distribution data samples. Based on scaffold split, we split the molecules in each dataset into training set, validation set, and test set by the ratio of 8:1:1. We run each method for 100 epochs on the training set in the training process and then select the epoch according to the validation set. The selected epoch is evaluated on the test set.



\textbf{GNN Architecture.}
We use the $\textit{AGGREGATE}$ function and $\textit{COMBINE}$ function defined in Graph Isomorphism Network (GIN) \cite{DBLP:conf/iclr/XuHLJ19}. To further improve the performance, residual connections \cite{DBLP:conf/cvpr/HeZRS16}, layer normalization \cite{DBLP:journals/corr/BaKH16}, and graph normalization \cite{DBLP:journals/corr/abs-2009-11746} are incorporated into GIN. Also, we use the average pooling as the $\textit{READOUT}$ function to obtain the graph representation.

\textbf{Hyper-parameters and Evaluation Metrics.}
We use Adam Optimizer\cite{DBLP:journals/corr/KingmaB14} with learning rate of 0.001 for all our models. For each dataset, we train the model with batch size of 32. As suggested by the MoleculeNet \cite{DBLP:journals/corr/WuRFGGPLP17},  we use the average ROC-AUC \cite{DBLP:journals/pr/Bradley97} as the evaluation metric for the 6 binary classification datasets. With respect to the regression datasets, for FreeSolv \cite{DBLP:journals/jcamd/MobleyG14}, ESOL \cite{DBLP:journals/jcisd/Delaney04}, and Lipo \cite{DBLP:journals/nar/GaultonBBCDHLMMAO12}, we use Root Mean Square Error (RMSE), and for QM7 \cite{blum2009970}, QM8 \cite{ramakrishnan2015electronic}, and QM9 \cite{DBLP:journals/jcisd/RuddigkeitDBR12}, we use Mean Average Error (MAE). We execute 4 independent runs for each method and report the mean and the standard deviation of the metrics.

\textbf{Baselines.}
We compare the proposed method with various competitive baselines. D-MPNN \cite{doi:10.1021/acs.jcim.9b00237} and AttentiveFP \cite{doi:10.1021/acs.jmedchem.9b00959} are the GNNs without pre-training, while N-Gram \cite{DBLP:conf/nips/LiuDL19}, PretrainGNN \cite{DBLP:conf/iclr/HuLGZLPL20}, and GROVER \cite{DBLP:conf/nips/RongBXX0HH20} are the methods with pre-training. N-Gram assembles the node embeddings in short walks in the graph to obtain the graph representation and then leverages Random Forest or XGBoost to predict the molecular properties. PretrainGNN implements several types of self-supervised learning tasks, among which we report the best result. GROVER integrates GNN into Transformer with the context prediction task and the functional motif prediction task, and we report the results of $\text{GROVER}_{\text{base}}$ and $\text{GROVER}_{\text{large}}$ with different network capacity.

More network and experimental details can be found in the Appendices.


\subsection{Overall Performance}

\begin{table}[t]
  \small
  \caption{Overall performance for molecular property prediction.}
  \centering
  \setlength\tabcolsep{1pt}{
  \begin{tabular}{c|ccc|ccc}
    \toprule
    \multicolumn{7}{c} {Regression (Lower is better)} \\
    \midrule
    & \multicolumn{3}{c|} {RMSE} & \multicolumn{3}{c} {MAE}  \\
    \midrule
    Dataset & ESOL & FreeSolv  & Lipo & QM7 & QM8 & QM9 \\
    \#Molecules & 1128 & 642  & 4200 & 6830 & 21786 & 133885  \\
    \#Prediction tasks & 1 & 1 & 1 & 1 & 12 & 12 \\
    \midrule
    D-MPNN \cite{doi:10.1021/acs.jcim.9b00237} & $1.050_{(0.008)}$  & $2.082_{(0.082)}$  & \cellcolor{gray!30}$0.683_{(0.016)}$ & $103.5_{(8.6)}$ & $0.0190_{(0.0001)}$ & $0.00814_{(0.00001)}$\\
    AttentiveFP \cite{doi:10.1021/acs.jmedchem.9b00959} & \cellcolor{gray!30}$0.877_{(0.029)}$ & \cellcolor{gray!30}$2.073_{(0.183)}$  & $0.721_{(0.001)}$  & \cellcolor{gray!30}$72.0_{(2.7)}$ & \cellcolor{gray!30}$0.0179_{(0.0001)}$ & \cellcolor{gray!30}$0.00812_{(0.00001)}$\\
    \midrule
    $\text{N-Gram}_{\text{RF}}$ \cite{DBLP:conf/nips/LiuDL19}  & $1.074_{(0.107)}$ & $2.688_{(0.085)}$ & $0.812_{(0.028)}$ & $92.8_{(4.0)}$ & $0.0236_{(0.0006)}$ & $0.01037_{(0.00016)}$ \\
    $\text{N-Gram}_{\text{XGB}}$ \cite{DBLP:conf/nips/LiuDL19} & $1.083_{(0.082)}$ & $5.061_{(0.744)}$ & $2.072_{(0.030)}$ & $81.9_{(1.9)}$& $0.0215_{(0.0005)}$& $0.00964_{(0.00031)}$ \\
    PretrainGNN \cite{DBLP:conf/iclr/HuLGZLPL20} & $1.100_{(0.006)}$ & $2.764_{(0.002)}$ & $0.739_{(0.003)}$ & $113.2_{(0.6)}$ & $0.0200_{(0.0001)}$ & $0.00922_{(0.00004)}$ \\
    $\text{GROVER}_{\text{base}}$ \cite{DBLP:conf/nips/RongBXX0HH20} & $0.983_{(0.090)}$ & $2.176_{(0.052)}$ & $0.817_{(0.008)}$ & $94.5_{(3.8)}$ & $0.0218_{(0.0004)}$ & $0.00984_{(0.00055)}$ \\
    $\text{GROVER}_{\text{large}}$ \cite{DBLP:conf/nips/RongBXX0HH20} & $0.895_{(0.017)}$ & $2.272_{(0.051)}$ & $0.823_{(0.010)}$ & $92.0_{(0.9)}$ & $0.0224_{(0.0003)}$ & $0.00986_{(0.00025)}$   \\
    \midrule
    ChemRL-GEM  & $\mathbf{0.798_{(0.029)}}$ & $\mathbf{1.877_{(0.094)}}$ & $\mathbf{0.660_{(0.008)}}$ & $\mathbf{58.9_{(0.8)}}$ & $\mathbf{0.0171_{(0.0001)}}$ & $\mathbf{0.00746_{(0.00001)}}$ \\
    \bottomrule
  \end{tabular}}
  
  \vspace*{5pt}
  \setlength\tabcolsep{1pt}{
  \begin{tabular}{c|cccccc|c}
    \toprule
    \multicolumn{8}{c}{Classification (Higher is better)} \\
    \midrule
    Dataset  & BACE & BBBP & ClinTox & SIDER & Tox21 & ToxCast   & Avg \\
    \#Molecules  & 1513  & 2039 & 1478 & 1427 & 7831 & 8575  & -  \\
    \#Prediction tasks & 1 & 1 & 2 & 27 & 12 & 617 & - \\
    \midrule
    D-MPNN \cite{doi:10.1021/acs.jcim.9b00237} & $0.809_{(0.006)}$ & \cellcolor{gray!30} $0.710_{(0.003)}$  & \cellcolor{gray!30}$0.906_{(0.006)}$ & $0.570_{(0.007)}$ & $0.759_{(0.007)}$ & $0.655_{(0.003)}$ & \cellcolor{gray!30}0.735\\
    AttentiveFP \cite{doi:10.1021/acs.jmedchem.9b00959} & $0.784_{(0.022)}$  & $0.643_{(0.018)}$ & $0.847_{(0.003)}$ & $0.606_{(0.032)}$  & $0.761_{(0.005)}$ & $0.637_{(0.002)}$ & 0.713 \\
    \midrule
    $\text{N-Gram}_{\text{RF}}$ \cite{DBLP:conf/nips/LiuDL19} & $0.779_{(0.015)}$ & $0.697_{(0.006)}$ & $0.775_{(0.040)}$ & \cellcolor{gray!30}$0.668_{(0.007)}$  & $0.743_{(0.004)}$ & -  & - \\
    $\text{N-Gram}_{\text{XGB}}$ \cite{DBLP:conf/nips/LiuDL19} & $0.791_{(0.013)}$  & $0.691_{(0.008)}$ & $0.875_{(0.027)}$ & $0.655_{(0.007)}$  & $0.758_{(0.009)}$  & -  & - \\
    PretrainGNN  \cite{DBLP:conf/iclr/HuLGZLPL20} & \cellcolor{gray!30}$0.845_{(0.007)}$   & $0.687_{(0.013)}$ & $0.726_{(0.015)}$ & $0.627_{(0.008)}$  & \cellcolor{gray!30}$0.781_{(0.006)}$ & \cellcolor{gray!30}$0.657_{(0.006)}$  &0.721 \\
    $\text{GROVER}_{\text{base}}$ \cite{DBLP:conf/nips/RongBXX0HH20} & $0.826_{(0.007)}$  & $0.700_{(0.001)}$ & $0.812_{(0.030)}$ & $0.648_{(0.006)}$ & $0.743_{(0.001)}$ & $0.654_{(0.004)}$  & 0.730 \\
    $\text{GROVER}_{\text{large}}$ \cite{DBLP:conf/nips/RongBXX0HH20} & $0.810_{(0.014)}$ & $0.695_{(0.001)}$ & $0.762_{(0.037)}$ & $0.654_{(0.001)}$ & $0.735_{(0.001)}$ & $0.653_{(0.005)}$ & 0.718 \\
    \midrule
    ChemRL-GEM & $\mathbf{0.856_{(0.011)}}$   & $\mathbf{0.724_{(0.004)}}$  & $0.901_{(0.013)}$ & $ \mathbf{{0.672_{(0.004)}}}$   & $\mathbf{0.781_{(0.001)}}$ & $\mathbf{0.692_{(0.004)}}$ & $\mathbf{0.771}$ \\
    \bottomrule
  \end{tabular}}
  \label{tbl:overall_performance}
  \vspace{-2em}
\end{table}

The overall performance of ChemRL-GEM along with other methods on the molecular property prediction benchmarks is summarized in Table~\ref{tbl:overall_performance}\footnote{Since N-Gram on ToxCast is too time-consuming, we are not able to finish in time but will be added later.}, where the SOTA results are shown in bold and the cells in gray indicate the previous SOTA results. The numbers in brackets are the standard deviation. From Table~\ref{tbl:overall_performance}, we have the following observations:
1) ChemRL-GEM achieves SOTA results on 11/12 datasets. On the regression tasks, ChemRL-GEM achieves an overall relative improvement of $8.8\%$ on average compared to the previous SOTA results in each dataset. While on the classification tasks, it achieves an overall relative improvement of $3.7\%$ on the average ROC-AUC compared to the previous SOTA result from D-MPNN. 2) Since the tasks of the regression datasets, such as the water solubility prediction in the ESOL dataset and the electronic properties prediction in the QM7 dataset, are much more correlated to the molecular geometries than the classification tasks, ChemRL-GEM achieves more considerable improvement on the regression datasets compared to the classification datasets. 3) ChemRL-GEM does not achieve SOTA results in the Clintox. It is possible that the Clintox dataset is highly unbalanced, with only 9 positive samples in the test set, which may cause unstable results. We also conduct experiments on other splitting methods, and ChemRL-GEM still achieves SOTA results. Please refer to the Appendix~\ref{appendix:exp} for more details.

\subsection{Ablation Studies of ChemRL-GEM}
\textbf{Contribution of GeoGNN.}
We investigate the effect of GeoGNN on regression datasets which are more related to the molecular geometries. GeoGNN is compared with multiple GNN architectures, including the commonly used GNN architectures, GIN \cite{DBLP:conf/iclr/XuHLJ19}, GAT \cite{DBLP:journals/corr/abs-1710-10903}, and GCN \cite{DBLP:journals/corr/KipfW16}, as well as the architectures specially designed for molecular representation, D-MPNN \cite{doi:10.1021/acs.jcim.9b00237}, AttentiveFP \cite{doi:10.1021/acs.jmedchem.9b00959}, and GTransformer \cite{DBLP:conf/nips/RongBXX0HH20}. From Table~\ref{tbl:geognn}, we can conclude that GeoGNN significantly outperforms other GNN architectures on all the regression datasets since GeoGNN utilizes a clever architecture that incorporates geometrical parameters even though the 3D coordinates of the atoms are simulated. The overall relative improvement is $7.9\%$ compared to the best results of previous methods.

Although the simulated 3D coordinates by RDKit are not accurate, they still provide the coarse information about the 3D conformation. Furthermore, we instead utilize the 3D coordinates of the molecules provided by QM9, which are much more accurate.  Surprisingly GeoGNN with the accurate 3D coordinates achieves an average MAE of $0.00652$, while GeoGNN with the coarse 3D coordinates achieves an average MAE of $0.00746$. Such improvement further demonstrates the increasing power of GeoGNN on learning molecular representations when providing more accurate 3D coordinates.

\begin{table}[t]
  \small
  \caption{Performance of different GNN architectures on the regression datasets.}
  \centering
  \setlength\tabcolsep{1pt}{
  \begin{tabular}{c|ccc|ccc}
    \toprule
    & \multicolumn{3}{c|} {RMSE} & \multicolumn{3}{c} {MAE}  \\
    \midrule
     Method & ESOL & FreeSolv & Lipo & QM7 & QM8 & QM9\\
    \midrule
    GIN \cite{DBLP:conf/iclr/XuHLJ19} & $1.067_{(0.051)}$ & $2.346_{(0.122)}$ & $0.757_{(0.022)}$ & $110.3_{(7.2)}$ & $0.0199_{(0.0002)}$ & $0.00886_{(0.00005)}$ \\
    GAT \cite{DBLP:journals/corr/abs-1710-10903} & $1.556_{(0.085)}$ & $3.559_{(0.050)}$ & $1.021_{(0.029)}$ & $103.0_{(4.4)}$ & $0.0224_{(0.0005)}$ & $0.01117_{(0.00018)}$  \\
    GCN \cite{DBLP:journals/corr/KipfW16} & $1.211_{(0.052)}$ & $3.174_{(0.308)}$  & $0.773_{(0.007)}$ & $100.0_{(3.8)}$ & $0.0203_{(0.0005)}$ & $0.00923_{(0.00019)}$  \\
    \midrule
    D-MPNN \cite{doi:10.1021/acs.jcim.9b00237}  & $1.050_{(0.008)}$ & $2.082_{(0.082)}$ & $0.683_{(0.016)}$ & $103.5_{(8.6)}$ & $0.0190_{(0.0001)}$ & $0.00814_{(0.00009)}$\\
    AttentiveFP \cite{doi:10.1021/acs.jmedchem.9b00959} & $0.877_{(0.029)}$ & $2.073_{(0.183)}$ & $0.721_{(0.001)}$  & $72.0_{(2.7)}$ & $0.0179_{(0.0001)}$ & $0.00812_{(0.00001)}$\\
    GTransformer \cite{DBLP:conf/nips/RongBXX0HH20}  & $2.298_{(0.118)}$ & $4.480_{(0.155)}$ & $1.112_{(0.029)}$ & $161.3_{(7.1)}$ & $0.0361_{(0.0008)}$ & $0.00923_{(0.00019)}$  \\
    \midrule
    GeoGNN & $\mathbf {0.832_{(0.010)}}$ & $\mathbf {1.857_{(0.071)}}$ & $\mathbf{0.666_{(0.015)}}$ & $\mathbf {59.0_{(3.4)}}$ & $\mathbf {0.0173_{(0.0004)}}$ & $\mathbf {0.00746_{(0.00003)}}$ \\
  \bottomrule
  \end{tabular}}
  \label{tbl:geognn}
  \vspace{-0.5em}
\end{table}

\begin{table}[t]
  \small
  \caption{Performance of different pre-training strategies on the regression datasets.}
  \centering
  \setlength\tabcolsep{1pt}{
  \begin{tabular}{c|ccc|ccc}
    \toprule
    & \multicolumn{3}{c|} {RMSE} & \multicolumn{3}{c} {MAE}  \\
    \midrule
     Pre-train Method & ESOL  & FreeSolv & Lipo & QM7 & QM8 & QM9\\
    \midrule
    w/o pre-train & $0.832_{(0.010)}$ & $1.857_{(0.071)}$ & $0.666_{(0.015)}$ & $59.0_{(3.4)}$ & $0.0173_{(0.0004)}$ & $0.00746_{(0.00003)}$ \\
    \midrule
    Context+Graph & $0.837_{(0.027)}$ & $1.982_{(0.098)}$ & $0.664_{(0.011)}$ & $72.1_{(2.3)}$ &$\mathbf{0.0171_{(0.0003)}}$ &$0.00748_{(0.00005)}$   \\
    Graph & $0.815_{(0.025)}$ & $1.950_{(0.069)}$ & $0.665_{(0.012)}$ & $63.1_{(2.8)}$ &$0.0174_{(0.0002)}$ &$0.00750_{(0.00001)}$   \\
    Geometry & $0.825_{(0.017)}$ & $\mathbf{1.701_{(0.147)}}$ & $\mathbf{0.660_{(0.021)}}$ & $\mathbf{58.2_{(0.5)}}$ & $\mathbf{0.0171_{(0.0001)}}$ & $\mathbf{0.00734_{(0.00003)}}$ \\
    Geometry+Graph & $\mathbf{0.798_{(0.029)}}$ & $1.876_{(0.094)}$ & $\mathbf{0.660_{(0.008)}}$ & $58.9_{(0.8)}$ & $\mathbf{0.0171_{(0.0001)}}$ & $0.00746_{(0.00001)}$  \\
  \bottomrule
  \end{tabular}}
  \label{tbl:self-supervised}
  \vspace{-2em}
\end{table}

\textbf{Contribution of Geometry-level Tasks.}
To study the effect of the proposed geometry-level self-supervised learning tasks, we apply different types of self-supervised learning tasks to pre-train GeoGNN on the regression datasets. In Table~\ref{tbl:self-supervised}, $\textit{w/o pre-training}$ denotes the GeoGNN network without pre-training, $\textit{Geometry}$ denotes our proposed geometry-level tasks, $\textit{Graph}$ denotes the graph-level task that predicts the molecular fingerprints, and $\textit{Context}$ \cite{DBLP:conf/nips/RongBXX0HH20} denotes a node-level task that predicts the atomic context. In general, the methods with geometry-level tasks are better than that without it. Furthermore, \textit{Geometry} performs better than \textit{Geometry+Graph} in the regression tasks, which may due to the weak connection between molecular fingerprints and the regression tasks.

\section{Related Work}
\subsection{Molecular Representation}
Current molecular representations can be categorized into three types: molecular fingerprints, sequence-based representations and graph-based representations.

\textbf{Molecular Fingerprints.} Molecular fingerprints, such as ECFP \cite{DBLP:journals/jcisd/RogersH10} and MACCS \cite{DBLP:journals/jcisd/DurantLHN02}, are commonly used for molecular representations by traditional machine learning methods \cite{cereto2015molecular,coley2017convolutional,DBLP:conf/nips/DuvenaudMABHAA15,DBLP:journals/corr/abs-2004-08919}, which encode a molecule into a sequence of bits according to the molecules' topological substructures. However, molecular fingerprints lack the ability to represent complex global structures, since they only focus on the local substructures.

\textbf{Sequence-based Representations.}
Some studies \cite{goh2018smiles2vec,DBLP:journals/corr/abs-2004-08919} take SMILES strings \cite{DBLP:journals/jcisd/Weininger88} that describe the molecules by strings as inputs, and leverage sequence-based models, such as Recurrent Neural Networks and Transformer \cite{DBLP:journals/corr/ZarembaSV14,vaswani2017attention}, to learn the molecular representations.
However, the same molecule could be represented by more than two SMILES strings, resulting in ambiguity of the representation. Besides, it is laborious for sequence-based representation to model some molecular topological structures, such as rings.

\textbf{Graph-based Representations.}
Many works \cite{DBLP:journals/corr/abs-2004-08919,DBLP:conf/nips/RongBXX0HH20,DBLP:journals/corr/abs-1909-00259,DBLP:conf/icdm/ShuiK20,DBLP:conf/icml/GilmerSRVD17} have showcased the great potential of graph neural networks on modeling molecules by taking each atom as a node and each chemical bond as an edge. For example, Attentive FP \cite{doi:10.1021/acs.jmedchem.9b00959} proposes to extend graph attention mechanism in order to learn aggregation weights. Furthermore, several works \cite{DBLP:conf/nips/SchuttKFCTM17,DBLP:journals/corr/abs-2102-07933} start to take the atomic distance into edge features to consider partial geometry information. However, they still lack the ability to model the full geometry information due to the shortage of traditional GNN architecture.

\subsection{Pre-training for Graph Neural Networks}
Self-supervised learning \cite{DBLP:conf/naacl/DevlinCLT19,DBLP:journals/corr/DoerschGE15,gidaris2018unsupervised,DBLP:journals/corr/abs-2006-03654,DBLP:journals/corr/abs-2012-11175} has achieved great success in natural language processing (NLP), computer vision (CV), and other domains, which trains unlabeled samples in a supervised manner to alleviate the over-fitting issue and improve data utilization efficiency. Recently, some studies \cite{DBLP:conf/iclr/HuLGZLPL20,DBLP:conf/nips/RongBXX0HH20} apply self-supervised learning methods to GNNs for molecular property prediction to overcome the insufficiency of the labeled samples. These works learn the molecular representation vectors by exploiting the node-level and graph-level tasks, where the node-level tasks learn the local domain knowledge by predicting the node properties and the graph-level tasks learn the global domain knowledge by predicting biological activities. Although existing self-supervised learning methods can boost the generalization ability, they neglect the spatial knowledge that is strongly related to the molecular properties.

\section{Conclusions and Future Work}
Efficient molecular representation learning is crucial for molecular property prediction. Existing works that apply GNNs and pre-training methods for molecular property prediction fail to fully utilize the molecular geometries described by bonds, bond angles, and other geometrical parameters. To this end, we design a geometry-based GNN for learning the atom-bond-angle relations that utilize the information of the bond angles by introducing a bond-angle graph on top of the atom-bond graph. 
Moreover, multiple geometry-level self-supervised learning methods are constructed to predict the molecular geometries to capture spatial knowledge. In order to verify the effectiveness of ChemRL-GEM, extensive experiments were conducted, comparing it with multiple competitive baselines. ChemRL-GEM significantly outperforms other methods on 12 benchmarks. In the future, other geometric parameters, such as torsional angles, will be considered to further boost the molecular representation capacity. We will also study the performance of applying molecular representation to other molecule-related problems, including predicting drug-target interaction and the interactions between molecules.

\clearpage
\appendix
\appendixpage

\section{Details of GeoGNN} \label{appendix:overall_geognn}
\subsection{GNN Architecture}

\begin{figure}[h]
\centering
  \includegraphics[width=1\textwidth]{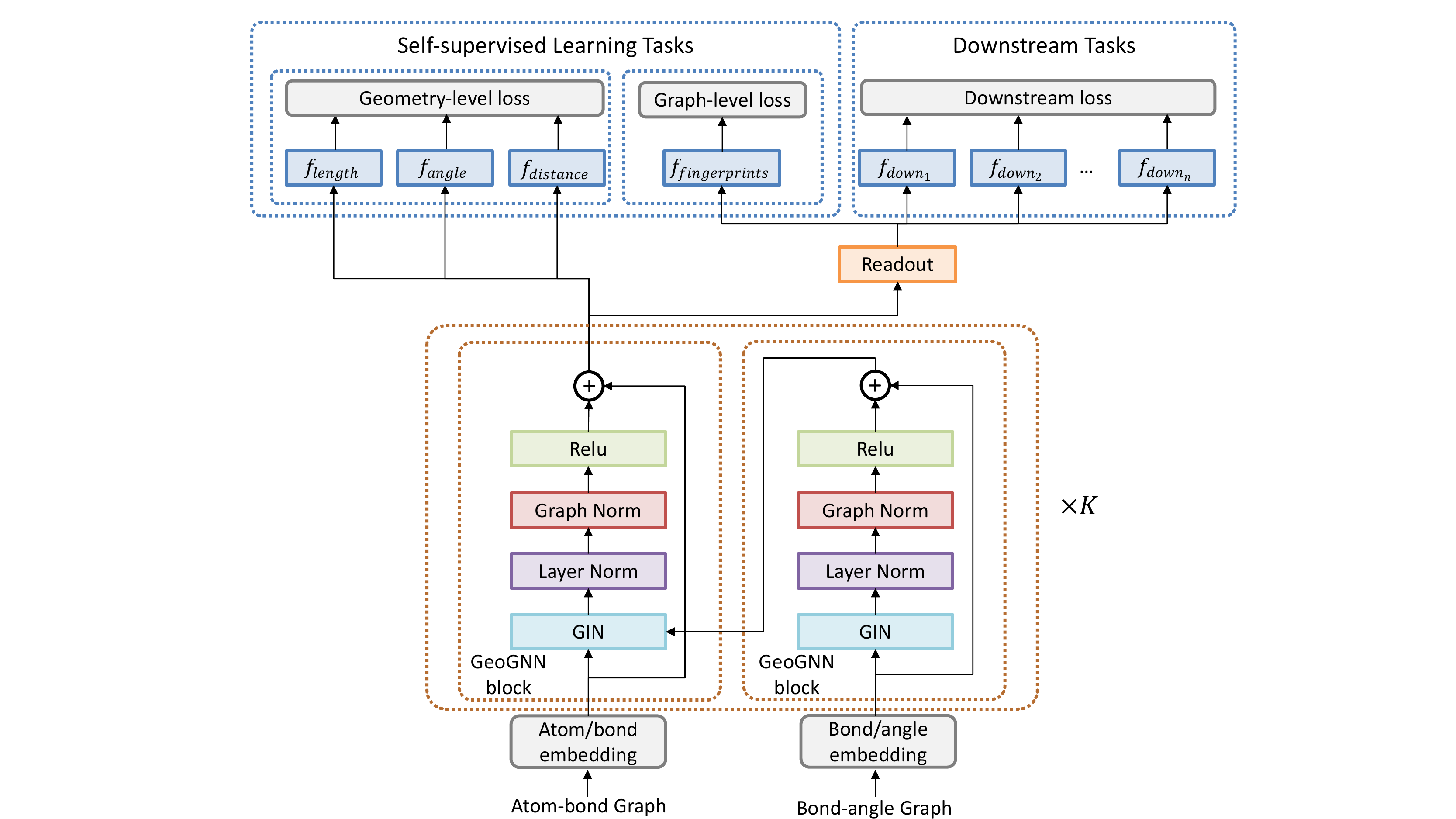}
  
\caption{Architecture of GeoGNN.}
\label{fig:geognn_block}
\end{figure}

As introduced in the main body, GeoGNN consists of two stacks of GeoGNN blocks, one for the atom-bond graph and the other for the bond-angle graph. The architecture of GeoGNN are shown in the bottom of Figure~\ref{fig:geognn_block}, where \textit{Layer Norm} \cite{DBLP:journals/corr/BaKH16}, \textit{Graph Size Norm} \cite{DBLP:journals/corr/abs-2009-11746} and \textit{Residual Connection} \cite{DBLP:conf/cvpr/HeZRS16} are popular tricks commonly used in GNNs. \textit{GIN} \cite{DBLP:conf/iclr/XuHLJ19}, a convolutional block for message passing, is utilized as the backbone of GeoGNN. In our experiments, the \textit{AGGREGATE} function and \textit{COMBINE} function in \textit{GIN} are defined as
\begin{equation*}
    \begin{split}
        \textit{AGGREGATE}_\textit{bond-angle}: a_{uv}^{(k)} =& \sum_{w \in \mathcal{N}(u)}(h_{uv}^{(k-1)} + h_{uw}^{(k-1)} + x_{wuv}) \\ 
        & + \sum_{w \in \mathcal{N}(v)}(h_{uv}^{(k-1)} + h_{vw}^{(k-1)} + x_{uvw}), \\
        \textit{COMBINE}_\textit{bond-angle}: h_{uv}^{(k)} =& \textit{MLP}( a_{uv}^{(k)}), \\
        \textit{AGGREGATE}_\textit{atom-bond}: a_u^{(k)} =& \sum_{v \in \mathcal{N}(u)}(h_u^{(k-1)} + h_v^{(k-1)} + h_{uv}^{(k-1)}), \\
        \textit{COMBINE}_\textit{atom-bond}: h_u^{(k)} =& \textit{MLP}( a_u^{(k)}).
    \end{split}
\end{equation*} 
where \textit{AGGREGATE} function summarizes the node features and the edge features, while \textit{COMBINE} function is a 2-layer Multi Layer Perceptron (MLP) with hidden size of 32. We use 8 GeoGNN blocks for atom-bond graph and bond-angle graph and the hidden size is set to be 32.

On top of GeoGNN, different headers are designed for different self-supervised tasks and downstream tasks while all the tasks share the same foundation, i.e., GeoGNN, as illustrated in Figure~\ref{fig:geognn_block}. More specifically, for the geometry-level tasks, the headers are applied upon the node representations $h_u^{(K)}$:
\begin{equation*}
    \begin{split}
        f_\textit{length}(h_u^{(K)}, h_v^{(K)}) = & \textit{MLP}(\textit{Concat}(h_u^{(K)}, h_v^{(K)})), \\
        f_\textit{angle}(h_u^{(K)}, h_v^{(K)}, h_w^{(K)}) = & \textit{MLP}(\textit{Concat}(h_u^{(K)}, h_v^{(K)}, h_w^{(K)})), \\
        f_\textit{distance}(h_u^{(K)}, h_v^{(K)}) = & \textit{MLP}(\textit{Concat}(h_u^{(K)}, h_v^{(K)})). \\
    \end{split}
\end{equation*} 
where \textit{MLP} is a 2-layer MLP network with hidde size of 256 and \textit{Concat} is the concatenation operation. While for the graph-level tasks and downstream tasks, the headers are applied upon the graph representations $h_G$:
\begin{equation*}
    \begin{split}
        f_\textit{fingerprints}(h_G) = & \textit{Linear}(h_G), \\
        f_{\textit{down}_n}(h_G) = & \textit{MLP}(h_G).
    \end{split}
\end{equation*}
where the \textit{MLP} is a 3-layer MLP network with hidden size of 128, and $f_{\textit{down}_1}$ to  $f_{\textit{down}_N}$ in Figure~\ref{fig:geognn_block} represent $N$ different downstream tasks.


\subsection{Input Features}

As stated in Section~\ref{Sec:geo_gnn}, the input features of GeoGNN can be categorized into three parts: the atom features, bond features, and bond angle features, shown in Table~\ref{tab:atom_features}. All these features are extracted by RDKit \cite{landrum2006rdkit} with the help of the Merck molecular force field function. Among them, bond lengths and bond angles are the continuous features, while the others are the discrete features. For the continuous features, we use the Radial Basis Functions \cite{buhmann2003radial} to expand each continuous value $x$ into a feature vector $e$ of dimension $M$:
\begin{equation}
    e_m(x) = \exp(-\gamma||x - \mu_m||^2),
\label{eq:rbf}
\end{equation}
where $\gamma$ controls the shape of the radial kernel, and we set $\gamma=10$. $\{\mu_m\}$ is a list of centers ranging from the minimum value to the maximum value of corresponding features with stride of 0.1. Besides, for the discrete features, they are converted into one-hot vectors according to their vocabulary size.

\begin{table}[t]
    \caption{Input features of GeoGNN}
    \small
    \label{tab:atom_features}
    \centering
    \setlength\tabcolsep{2pt}{
    \begin{tabular}{c|ccc}
        \toprule
        \multicolumn{1}{c|}{Feature type} & Feature & Description & Size \\
        \midrule
        \multirow{4}[8]{*}{atom} & atom type & type of atom (e.g., C, N, O), by atomic number (one-hot) & 119 \\
          &  aromaticity & whether the atom is part of an aromatic system (one-hot) & 2 \\
         & formal charge & electrical charge (one-hot) & 16 \\
         & chirality tag & CW, CCW, unspecified or other (ont-hot) & 4 \\
        &  degree & number of covalent bonds (one-hot) & 11 \\
         &  number of hydrogens & number of bonded hydrogen atoms (one-hot) & 9 \\
     &  hybridization & sp, sp\textsuperscript{2}, sp\textsuperscript{3}, sp\textsuperscript{3}d, or sp\textsuperscript{3}d\textsuperscript{2} (one-hot) & 5 \\
        \midrule
         \multirow{3}[6]{*}{bond} & bond dir & begin dash, begin wedge, etc. (one-hot) & 7 \\
        &  bond type & single, double, triple or aromatic (one-hot) & 4 \\
        & in ring & whether the bond is part of a ring (one-hot) & 2 \\
        & bond length &  bond length (float) & - \\
        \midrule
       {bond angle} & bond angle &  bond angle (float) & - \\
        \bottomrule
    \end{tabular}}
\end{table}

\section{Training and Test Processes for ChemRL-GEM}
The training and test processes of ChemRL-GEM is shown in Algorithm~\ref{alg:gem}.
\begin{itemize}
    \item Input. We introduce a pre-training dataset $\textbf{D}_{pre} = \{(G, H)\}$, which contains a great number of molecules and each molecule is represented by $(G, H)$. We divide each downstream dataset into train dataset $\textbf{D}_{down}^{train} = \{(G, H, y)\}$, validation dataset $\textbf{D}_{down}^{valid} = \{(G, H, y)\}$, and test dataset $\textbf{D}_{down}^{test} = \{(G, H, y)\}$, where $y$ denotes the label for molecule $(G, H)$. The training dataset $\textbf{D}_{down}^{train}$ is used to fine-tune the parameters, the validation dataset $\textbf{D}_{down}^{valid}$ is used to select the best epoch, and the test dataset $\textbf{D}_{down}^{test}$ is used to evaluate the performance. The pre-training dataset ${\textbf{D}}_{pre}$, downstream datasets $\textbf{D}_{down}^{train}$, $\textbf{D}_{down}^{valid}$, and $\textbf{D}_{down}^{test}$ are taken as the input of the algorithm.
    \item Model. GeoGNN is denoted as a function $F_{\Theta}(G, H)$, where $\Theta$ represents the set of model parameters. The model parameters are first pre-trained by the self-supervised learning tasks and then fine-tuned by the downstream tasks.
    \item Optimization. We combine Equation~\ref{eq:bond} and Equation~\ref{eq:distance_matrix} to obtain the overall loss function for pre-training: $L_{pre}(G,H)=L_{length}(\mathcal{E}) + L_{angle}(\mathcal{A}) + L_{distance}(\mathcal{V})$. $L_{down}(G,H, y)$ is introduced to denote the objective function of a downstream task.
\end{itemize} 

\begin{algorithm}
	\caption{ChemRL-GEM} 
	\begin{algorithmic}[1]
	    \Require pre-training dataset ${\textbf{D}}_{pre}$, downstream datasets $\textbf{D}_{down}^{train}$, $\textbf{D}_{down}^{valid}$, $\textbf{D}_{down}^{test}$
		\For {$epoch =1,2,\ldots$} \Comment{Pre-training}
			\For {$(G, H) \in \textbf{D}_{pre}$}
				\State Optimize $\Theta$ w.r.t. $L_{pre}(G,H)$
			\EndFor
		\EndFor
		\For {$epoch =1,2,\ldots$} \Comment{Fine-tuning}
			\For {$(G, H, y) \in \textbf{D}_{down}^{train}$}
				\State Fine-tune $\Theta$ w.r.t. $L_{down}(G,H,y)$
			\EndFor
			\For {$(G, H, y) \in \textbf{D}_{down}^{valid}$}
			    \State Evaluate $F_{\Theta}(G, H)$
			\EndFor
		\EndFor
	\State Select the best epoch w.r.t. the evaluation results on $\textbf{D}_{down}^{valid}$
	\For {$(G, H) \in \textbf{D}_{down}^{test}$} \Comment{Test}
		\State Evaluate $F_{\Theta}(G, H)$
	\EndFor
	\end{algorithmic} 
	\label{alg:gem}
\end{algorithm}

\section{Details of Experimental Settings}\label{appendix:exp}
\subsection{Dataset Description}

We select 6 molecular regression datasets and 6 molecular classification datasets from MoleculeNet \cite{DBLP:journals/corr/WuRFGGPLP17} as the benchmarks. These benchmarks can be categorized into the physical chemistry, quantum mechanics, biophysics, and physiology, as shown in Table~\ref{tab:dataset_info}. The benchmarks of physical chemistry and quantum mechanics are more related to molecular geometries compared to those of biophysics and physiology.

\begin{table}[h]
\small
    \caption{Dataset details}
    \label{tab:dataset_info}
    \centering
\begin{tabular}{c|c|cccc}
\toprule
\multicolumn{1}{c|}{{Task Type}} & \multicolumn{1}{c}{{Metric}} & \multicolumn{1}{|c}{{Category}} & \multicolumn{1}{c}{{Dataset}} & \multicolumn{1}{c}{{\# Tasks}} & \multicolumn{1}{c}{{\# Compounds}}  \\
\midrule
\multirow{4}[8]{*}{Regression}  &  \multirow{2}[4]{*}{RMSE} & \multirow{2}[4]{*}{Physical chemistry}  & ESOL  & 1     & 1,128  \\
   &     &      & FreeSolv & 1     & 642   \\
    &      &       & Lipophilicity & 1     & 4,200\\
\cline{2-6}   & \multirow{2}[4]{*}{MAE}   & \multirow{2}[4]{*}{Quantum mechanics} &  QM7   & 1     & 6,830   \\
   &     &       & QM8   & 12    & 21,786 \\
   &     &       & QM9   & 12    & 133,885 \\
\midrule
\multirow{4}[8]{*}{Classification} & \multirow{4}[8]{*}{ROC-AUC} &  Biophysics   & BACE  & 1     & 1,513   \\
\cline{3-6}   &     & \multirow{4}[8]{*}{Physiology} & BBBP  & 1     & 2,039  \\
     &  &       & ClinTox & 2     & 1,478   \\
     &  &       & SIDER & 27    & 1,427  \\
     &  &       & Tox21 & 12    & 7,831   \\
     &  &       & ToxCast & 617   & 8,575   \\

\bottomrule
\end{tabular}
\end{table}

The descriptions of various benchmarks are listed as follows:
\begin{itemize}
    \item \texttt{ESOL} \cite{DBLP:journals/jcisd/Delaney04} contains water solubility data (log solubility in mols per liter) for common organic small molecules. It is a standard dataset that is widely used to estimate solubility directly.
    \item  \texttt{FreeSolv} \cite{DBLP:journals/jcamd/MobleyG14} contains the experimental values of free energy of hydration of small molecules in water. The values are all obtained through molecular dynamics simulations. 
     \item \texttt{Lipophilicity} \cite{DBLP:journals/nar/GaultonBBCDHLMMAO12} is collected from the ChEMBL database \cite{DBLP:journals/nar/GaultonBBCDHLMMAO12}, containing the experimental results of the octanol or water partition coefficient, which reflects the solubility of the molecules. 
    \item \texttt{QM7} \cite{blum2009970} is a subset of GDB-13, which provides information about the spatial structures of the molecules. It records various electronic properties that are stable and synthetically obtainable, such as HOMO and LUMO determined by \textit{ab-initio} density function theory (DFT), and atomization energy.
    \item  \texttt{QM8} \cite{ramakrishnan2015electronic} uses a variety of quantum mechanics methods to calculate the electronic spectrum and excited state energy of small molecules.
    \item  \texttt{QM9}\footnote{Since the ranges of different targets in QM9 vary largely, we only take the ``homo'', ``lumo'', and ``gap'' targets to calculate the average MAE in our experiments.} \cite{DBLP:journals/jcisd/RuddigkeitDBR12} provides multiple data information on geometry, energy, electronic and thermodynamic properties of small molecules calculated by DFT.
    \item \texttt{BACE} \cite{subramanian2016computational} records molecules with 2D structures and properties, which provides the qualitative (binary label) binding results on inhibitors of human $\beta$-secretase 1.
    \item \texttt{BBBP} \cite{DBLP:journals/jcisd/MartinsTPF12} is a dataset that contains molecules with measured permeaility property of penetrating the blood-brain barrier.
    \item \texttt{ClinTox} \cite{gayvert2016data} includes drugs approved by the FDA and those that have failed clinical trials for toxicity reasons. 
    \item \texttt{SIDER} \cite{DBLP:journals/nar/KuhnLJB16} is a database of marketed drugs and adverse drug reactions (ADR), grouped into 27 system organ classes.
    \item \texttt{Tox21} \cite{huang2017editorial} measures the toxicity of compounds on 12 different targets, including nuclear receptors and stress response pathways. It has been used in the 2014 Tox21 Data Challenge as a public database. 
    \item \texttt{ToxCast} \cite{richard2016toxcast} includes toxicology results of thousands of molecules. It provides multiple toxicity labels by running high-throughput screening experiments on a large library of chemicals.
\end{itemize}

\subsection{Dataset Splitting Methods} 

There are several popular splitting methods used in molecular property prediction datasets, including random splitting, scaffold splitting \cite{bemis1996properties} and random scaffold splitting. Scaffold splitting and random scaffold splitting offer more challenging yet realistic ways of splitting by keeping the molecules with the same scaffold in either the train, validation, or the test set. Since the random scaffold splitting introduces more randomness into the evaluation of different methods, we adopt the scaffold splitting in the experiments of our main body. At the same time, we additionally conduct experiments using the random scaffold splitting on the classification datasets following the same experimental settings used in GROVER \cite{DBLP:conf/nips/RongBXX0HH20}, and again ChemRL-GEM achieves the SOTA results with an overall relative improvement of $2.0\%$ compared to the previous SOTA results on all the datasets, as shown in Table~\ref{tab:overall_performance_random_scaffold}. Note that, the results of the baselines are directly copied from \cite{DBLP:conf/nips/RongBXX0HH20}.

\begin{table}[h]
  \caption{The performance of classification datasets under random scaffold splitting}
  \centering
  \small
  \setlength\tabcolsep{1pt}{
  \begin{tabular}{c|cccccc|c}
    \toprule
    \multicolumn{8}{c}{Classification (Higher is better)} \\
    \midrule
    Dataset & BACE & BBBP & ClinTox & SIDER   & Tox21 & ToxCast  & Avg \\
    \midrule
    D-MPNN \cite{doi:10.1021/acs.jcim.9b00237} & $0.852_{(0.053)}$ & $0.919_{(0.030)}$ & $0.897_{(0.040)}$ & $0.632_{(0.023)}$   & $0.826_{(0.023)}$ &  $0.718_{(0.011)}$ & 0.807\\
    AttentiveFP \cite{doi:10.1021/acs.jmedchem.9b00959} &  $0.863_{(0.015)}$ & $0.908_{(0.050)}$  & $0.933_{(0.020)}$  & $0.605_{(0.060)}$ & $0.807_{(0.020)}$ & $0.579_{(0.001)}$ & 0.783\\
    \midrule
    $\text{N-Gram}_{\text{XGB}}$ \cite{DBLP:conf/nips/LiuDL19} &  $0.876_{(0.035)}$ & $0.912_{(0.013)}$ & $0.855_{(0.037)}$ & $0.632_{(0.005)}$  & $0.769_{(0.027)}$ & - & -\\
    PretrainGNN \cite{DBLP:conf/iclr/HuLGZLPL20} & $0.851_{(0.027)}$  & $0.915_{(0.040)}$  & $0.762_{(0.058)}$ & $0.614_{(0.006)}$ & $0.811_{(0.015)}$ & $0.714_{(0.019)}$ & 0.778\\
    $\text{GROVER}_{\text{base}}$ \cite{DBLP:conf/nips/RongBXX0HH20} & $0.878_{(0.016)}$  & $0.936_{(0.008)}$ & $0.925_{(0.013)}$  & $0.656_{(0.006)}$ & $0.819_{(0.020)}$ & $0.723_{(0.010)}$ & 0.823\\
    $\text{GROVER}_{\text{large}}$ \cite{DBLP:conf/nips/RongBXX0HH20} & \cellcolor{gray!30}$0.894_{(0.028)}$ & \cellcolor{gray!30}$0.940_{(0.019)}$  & \cellcolor{gray!30}${0.944}_{(0.021)}$ & \cellcolor{gray!30}${0.658}_{(0.023)}$ & \cellcolor{gray!30}${0.831}_{(0.025)}$ & \cellcolor{gray!30}${0.737}_{(0.010)}$ & \cellcolor{gray!30}0.834\\
    \midrule
    ChemRL-GEM & $\mathbf{0.925_{(0.010)}}$ & $\mathbf{0.953_{(0.007)}}$ & $\mathbf{0.977_{(0.019)}}$ & $\mathbf{0.663_{(0.014)}}$   & $\mathbf{0.849_{(0.003)}}$ & $\mathbf{0.742_{(0.004)}}$ & $\mathbf{0.852}$ \\
    \bottomrule
  \end{tabular}}
  \label{tab:overall_performance_random_scaffold}
  \vspace{-1.5em}
\end{table}

One more thing to mention is that we inspect several public implementations of baselines on scaffold calculation and find that some of them consider the chirality of molecules and some do not. Such inconsistence results in a big difference in the splitting train, validation, and test sets. To make the comparison fair, we use the scaffold calculation that considers molecular chirality and apply it to all the experiments. Furthermore, to ensure the data alignment, we summarize the statistics of the labels on each benchmarks. For the regression tasks, we summarize the minimum, maximum, and mean values of the labels, as shown in Figure~\ref{tab:statistics_regression}, while for the classification tasks, we summarize the ratios of positive and negative samples as shown in Figure~\ref{tab:statistics_classification}. Since there are \textit{Nan} values in Tox21 and ToxCast, the positive ratio and the negative ratio can not be summed up to be 1 in these two datasets.

\begin{table}[h]
    \small
    \caption{Statistics of label values in 6 regression datasets}
    \label{tab:statistics_regression}
    \centering
    \setlength\tabcolsep{2pt}{
    \begin{tabular}{c|ccc|ccc|ccc}
        \toprule
        \multicolumn{1}{c|} {Dataset}  & \multicolumn{3}{c|} {Train Set} &  \multicolumn{3}{c|} {Validation Set}  &\multicolumn{3}{c} {Test Set} \\
        \midrule
        {Label value} & Min & Max & Mean  & Min & Max & Mean  & Min & Max & Mean \\
        \midrule
        ESOL & -1.16e+1 & 1.58e+0 & -2.87e+0 & -9.33e+0 & 1.10e+0 & -3.77e+0 & -8.80e+0 & 1.07e+0 & -3.80e+0 \\
        FreeSolv & -2.36e+1 & 3.16e+0 & -3.26e+0 & -2.55e+1 & 2.55e+0 & -6.05e+0 & -1.81e+1 & 3.43e+0 & -5.88e+0 \\
        Lipo & -1.50e+0 & 4.50e+0 & 2.16e+0 & -1.10e+0 & 4.49e+0 & 2.20e+0  & -1.30e+0 & 4.50e+0 & 2.36e+0 \\
        QM7 & -2.19e+3 & -4.05e+2 & -1.55e+3 & -2.19e+3 & -1.00e+3 & -1.55e+3  & -2.06e+3 & -8.26e+2 & -1.48e+3 \\
        QM8 & -3.00e-3  & 7.11e-1 & 1.31e-1 & -4.50e-7 & 5.86e-1 & 1.28e-1  & -2.43e-5 & 5.41e-1 & 1.32e-1 \\
        QM9 & -4.29e-1 & 6.22e-1 & 7.00e-3 & -3.15e-1 & 3.75e-1 & 9.00e-3  & -3.67e-1 & 3.88e-1 & 6.00e-3 \\
        
        \bottomrule
    \end{tabular}}
    \vspace{-1.5em}
\end{table}

\begin{table}[h]
    \caption{Ratios of positive and negative samples in 6 classification datasets}
    \label{tab:statistics_classification}
    \small
    \centering
    \begin{tabular}{c|cc|cc|cc}
        \toprule
        \multicolumn{1}{c|} {Dataset}  & \multicolumn{2}{c|} {Train Set} &  \multicolumn{2}{c|} {Validation Set}  &\multicolumn{2}{c} {Test Set} \\
        \midrule
        {Ratio} & Positive  & Negative  & Positive  & Negative  & Positive  & Negative  \\
        \midrule
        BACE & 0.3967 & 0.6033 & 0.8609 & 0.1391 & 0.5329 & 0.4671 \\
        BBBP & 0.8406 & 0.1594 & 0.3971 & 0.6029 & 0.5294 & 0.4706 \\
        ClinTox & 0.5068 & 0.4932 & 0.5034 & 0.4966 & 0.5034 & 0.4966 \\
        SIDER & 0.5642 & 0.4358 & 0.5918 & 0.4082 & 0.5706 & 0.4294 \\
        Tox21 & 0.0599 & 0.7880 & 0.0731 & 0.6872 & 0.0717 & 0.6796 \\
        ToxCast & 0.0230 & 0.2766 & 0.0232 & 0.2104 & 0.0324 & 0.2355 \\
        \bottomrule
    \end{tabular}
    \vspace{-1.5em}
\end{table}

\subsection{Pre-training and Downstream Fine-tuning}

On the pre-training stage, we utilize the distributed training with 8 GPUs and $batch=512$ for each GPU. We set the dropout rate as 0.2, the vocabulary size for discretizing atomic distance as 30, and the mask ratio as 0.15. Adam optimizer \cite{DBLP:journals/corr/KingmaB14} with learning rate of $0.005$ is utilized and we train 20 epochs for each pre-training method.

On the downstream fine-tuning stage, we utilize a single card to train each dataset. Different batch sizes are selected for different datasets: $batch=256$ for the large-size datasets QM8 and QM9; $batch=128$ for the medium-size datasets Tox21 and ToxCast; $batch=32$ for all the other datasets. We use Adam optimizer and train 100 epochs for each model. As the downstream tasks are sensitive to hyper-parameters, we apply a grid search on the dropout rate and the learning rate. For the dropout rate, we search $\{0.1, 0.2,0.5\}$. For the learning rate, we consider GeoGNN body and the downstream headers separately, where we search body-header learning rate pairs: $\{(0.001,0.001),(0.004,0.004),(0.0001,0.001)\}$.

\newpage
\small
\bibliographystyle{plain}
\bibliography{references}
\end{document}